\documentclass[letterpaper]{article} 
\usepackage{aaai2026}  
\usepackage{times}  
\usepackage{helvet}  
\usepackage{courier}  
\usepackage[hyphens]{url}  
\usepackage{graphicx} 
\usepackage{natbib}  
\usepackage{caption} 
\usepackage{algorithm}
\usepackage{algorithmic}

\definecolor{myPink}{RGB}{255 ,105 ,180}
\usepackage{graphicx}
\usepackage{amsmath}
\usepackage{amssymb}
\usepackage{booktabs}
\usepackage{makecell}
\usepackage{colortbl}
\usepackage{xcolor}
\usepackage{booktabs}

\usepackage{pifont}
\newcommand{\xmark}{\ding{55}} 
\newcommand{\cmark}{\ding{51}} 
\usepackage{diagbox}

\usepackage{newfloat}
\usepackage{listings}
\DeclareCaptionStyle{ruled}{labelfont=normalfont,labelsep=colon,strut=off} 
\floatstyle{ruled}
\newfloat{listing}{tb}{lst}{}
\floatname{listing}{Listing}
\title{MotionCharacter: Fine-Grained Motion Controllable Human Video Generation}
\author{
    Haopeng Fang\textsuperscript{\rm 1}\thanks{Work was done during an internship at Meituan.},
    Di Qiu\textsuperscript{\rm 2}\thanks{Corresponding authors.},
    Binjie Mao\textsuperscript{\rm 2},
    He Tang\textsuperscript{\rm 1}\footnotemark[2]
}
\affiliations{
    \textsuperscript{\rm 1}School of Software Engineering, Huazhong University of Science and Technology, Wuhan, China\\
    \textsuperscript{\rm 2}Meituan, Beijing, China\\
    haopengfang@hust.edu.cn, 
    \{qiudihk, binjiemao\}@gmail.com, 
    hetang@hust.edu.cn
}

\usepackage{bibentry}

\begin{document}

\maketitle

\begin{abstract}
Recent advancements in personalized Text-to-Video (T2V) generation have made significant strides in synthesizing character-specific content. However, these methods face a critical limitation: the inability to perform fine-grained control over motion intensity. This limitation stems from an inherent entanglement of action semantics and their corresponding magnitudes within coarse textual descriptions, hindering the generation of nuanced human videos and limiting their applicability in scenarios demanding high precision, such as animating virtual avatars or synthesizing subtle micro-expressions. Furthermore, existing approaches often struggle to preserve high identity fidelity when other attributes are modified. To address these challenges, we introduce \textit{MotionCharacter}, a framework for high-fidelity human video generation with precise motion control. At its core, \textit{MotionCharacter} explicitly decouples motion into two independently controllable components: action type and motion intensity. This is achieved through two key technical contributions: (1) a Motion Control Module that leverages textual phrases to specify the action type and a quantifiable metric derived from optical flow to modulate its intensity, guided by a region-aware loss that localizes motion to relevant subject areas; and (2) an ID Content Insertion Module coupled with an ID-Consistency loss to ensure robust identity preservation during dynamic motions. To facilitate training for such fine-grained control, we also curate Human-Motion, a new large-scale dataset with detailed annotations for both motion and facial features. Extensive experiments demonstrate that \textit{MotionCharacter} achieves substantial improvements over existing methods. Our framework excels in generating videos that are not only identity-consistent but also precisely adhere to specified motion types and intensities. 
\end{abstract}

\begin{links}
    \link{Project Page}{https://motioncharacter.github.io/}
\end{links}

\section{Introduction}
\label{sec:intro}

\begin{figure}[ht!]
    \centering
    \includegraphics[width=\columnwidth]{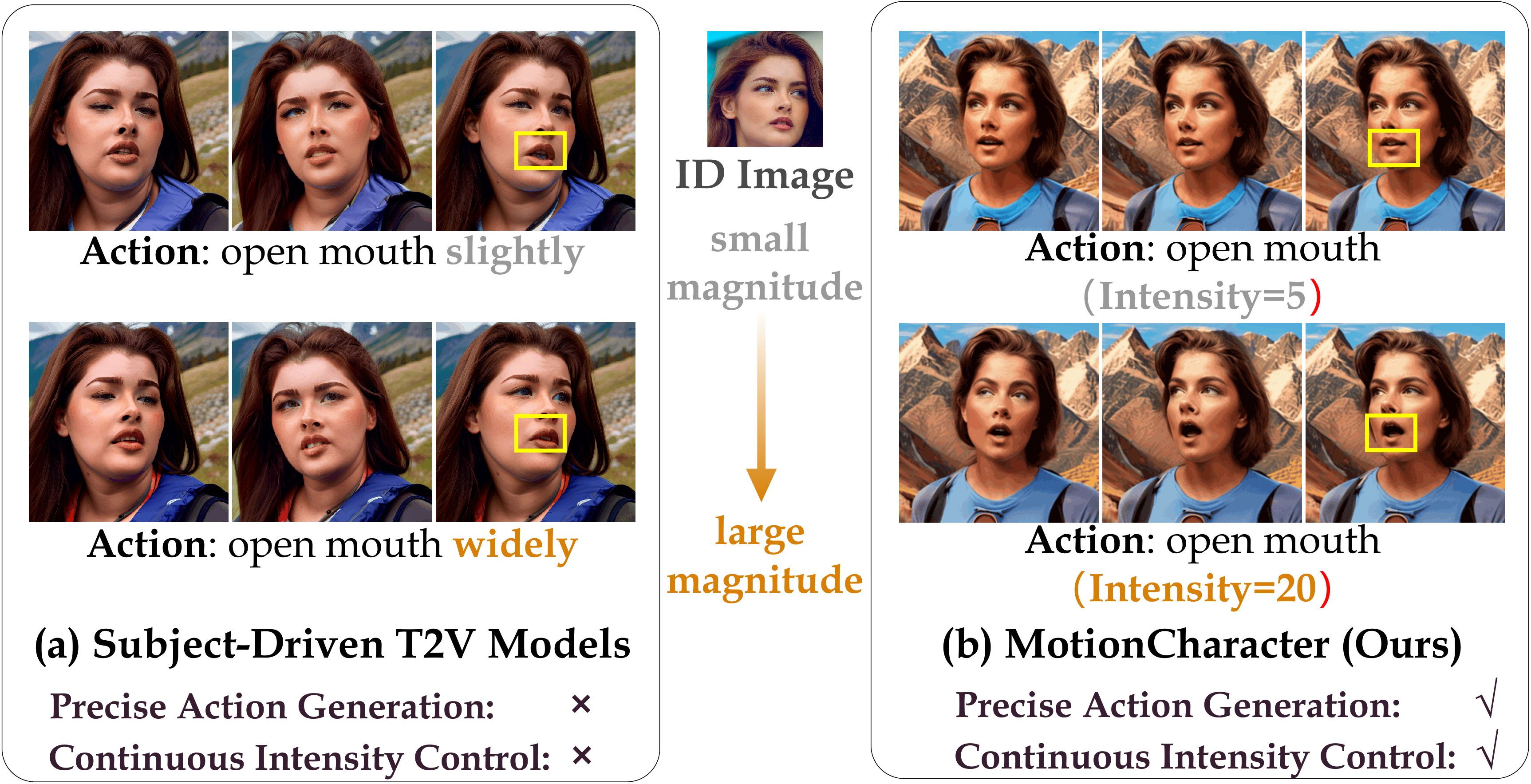}
    \caption{Comparison of subject-driven T2V methods and our proposed \emph{MotionCharacter} framework. Existing approaches specify coarse actions (e.g., ``open mouth’’) but struggle to capture nuanced motion magnitude (e.g., ``slightly’’ vs. ``widely’’). In contrast, \emph{MotionCharacter} decouples action type and motion intensity, enabling fine-grained, continuous control over human motion while preserving subject fidelity.}
    \label{fig:fig1}
\end{figure}

High-quality, personalized, and controllable human video generation has gained significant traction, with applications spanning social media, film production, virtual avatars, and personalized content creation. Recent pioneering advancements in text-driven video generation models~\cite{ho2022video, guo2023animatediff, ho2022imagen, wang2023modelscope, wang2023lavie, zhou2022magicvideo, chen2024videocrafter2} have driven substantial progress in this field. In particular, subject-driven Text-to-Video (T2V) generation models~\cite{jiang2024videobooth, wei2024dreamvideo, ma2024magic, wu2024customcrafter, he2024id} have made strides in producing high-quality videos that faithfully depict specific individuals. 

However, existing subject-driven T2V models still face a significant limitation: a lack of fine-grained control over motion. This deficiency curtails their applicability in scenarios demanding high precision,  such as animating digital humans with precise expressions, synthesizing subtle micro-expressions for psychological analysis, or creating realistic virtual avatars that respond dynamically to user input. 

The core of this problem lies in the entanglement of action semantics and motion intensity within textual prompts. As shown in Fig.~\ref{fig:fig1}(a), current subject-driven T2V models~\cite{he2024id, guo2023animatediff} allow users to specify human actions using action phrases like ``open mouth”. However, these phrases provide only a coarse description and fall short in precisely controlling motion intensity. Even when employing more nuanced phrases (e.g., ``open mouth slightly", ``open mouth widely"), the generated results often lack the intended subtleties because language captures actions in a discrete manner, whereas motion intensity is inherently continuous. This forces the model to guess the intended magnitude, leading to unpredictable and uncontrollable results. 

To break this entanglement and enable fine-grained controllability, we introduce \textit{MotionCharacter}, a framework that explicitly decouples motion into two components: action type and motion intensity. To achieve this, we first propose a Motion Control Module. This module uses simple action phrases (e.g., ``smile") to define the type of motion, while a separate, continuous signal derived from optical flow modulates its intensity. This allows a user to specify an action with text and then fine-tune its magnitude precisely, for instance, via a simple slider. To further enhance motion dynamics and the quality of facial transitions, this module is enhanced by a region-aware loss that directs the model’s attention to critical facial regions like the lips and eyes, preventing distortion during motion.

Moreover, as motions become more dynamic and expressive, maintaining the subject's identity becomes a significant challenge. To address this, we introduce an ID Preservation Module that injects detailed facial features into the generation process. This is reinforced by a powerful ID consistency loss to ensure that the character's core identity remains stable and detailed, even during large and complex movements. 

Finally, while several human video datasets~\cite{zhu2022celebv, yu2023celebv} exist, they predominantly focus on capturing basic emotional expressions or broad action categories, lacking the granular annotations needed for fine-grained motion control. Also, these datasets are limited in scale and contain unfiltered multi-subject and low-quality scenes. This limitation creates a significant bottleneck in training models capable of nuanced motion generation. To address this gap, we introduce Human-Motion, a large-scale dataset of 106,292 video clips. We use Large Multimodal Models (LMMs) and optical flow estimation to generate detailed annotations for motion type and intensity, aiming to facilitate research on fine-grained human video generation.

Through extensive experiments, we present qualitative, quantitative, and user study results that validate the performance of our method in terms of identity consistency and adherence to motion instructions. In summary, our contributions are as follows:

\begin{enumerate}
\item We propose \textit{MotionCharacter}, a framework for personalized human video generation that decouples motion control into action type and intensity, enabling fine-grained and continuous motion adjustments.
\item We introduce a Motion Control Module that leverages text for action semantics and a continuous signal for intensity, enhanced by a region-aware loss to ensure high-quality dynamics. We also propose an ID Preservation Module with a dedicated ID-Consistency loss to maintain robust identity fidelity.
\item We curate Human-Motion, a large-scale and high-quality dataset of 106,292 video clips with detailed annotations for motion and identity, created specifically to facilitate research on controllable human video generation.
\end{enumerate}

\section{Related Work}
\label{sec:2_relatedwork}
\noindent\textbf{Text-to-Video Diffusion Model.}
Recent advancements in diffusion models~\cite{ho2020denoising,song2020score,rombach2022high} have established them as a leading approach in Text-to-Video generation. The Video Diffusion Model~\cite{ho2022video} pioneered a space-time-factored U-Net for unconditional video generation in pixel space. Building on this, AnimateDiff~\cite{guo2023animatediff} integrated a motion module into the Stable Diffusion framework to generate videos from text prompts. Later works, such as Imagen Video~\cite{ho2022imagen} and Make-a-Video~\cite{singer2022make}, adopted sequential models for pixel-space T2V generation. To overcome challenges with high-dimensional video data, latent diffusion models~\cite{blattmann2023align} operate in the latent space of an auto-encoder, inspiring further developments including ModelScope~\cite{wang2023modelscope}, LAVIE~\cite{wang2023lavie}, MagicVideo~\cite{zhou2022magicvideo}, and VideoCrafter~\cite{chen2023videocrafter1, chen2024videocrafter2}.

\noindent\textbf{Subject-Driven Image and Video Generation.}
Subject-driven generation aims to synthesize content while preserving specific subject characteristics. In the image domain, recent works like IP-Adapter~\cite{ye2023ip}, PhotoMaker~\cite{li2024photomaker}, and InstantID~\cite{wang2024instantid} have achieved efficient identity preservation through embedding-based approaches, eliminating the need for subject-specific training required by methods like DreamBooth~\cite{ruiz2023dreambooth}. For video generation, works such as VideoBooth~\cite{jiang2024videobooth}, DreamVideo~\cite{wei2024dreamvideo}, and CustomCrafter~\cite{wu2024customcrafter} have explored learning-based frameworks to combine visual identity with motion dynamics. While ID-Animator~\cite{he2024id} demonstrates zero-shot capabilities, it lacks fine-grained control over motion intensity. Our \textit{MotionCharacter} addresses these limitations by enabling control of both appearance and motion without requiring retraining during inference.

\begin{figure*}[ht]
    \centering
    \includegraphics[width=\textwidth]{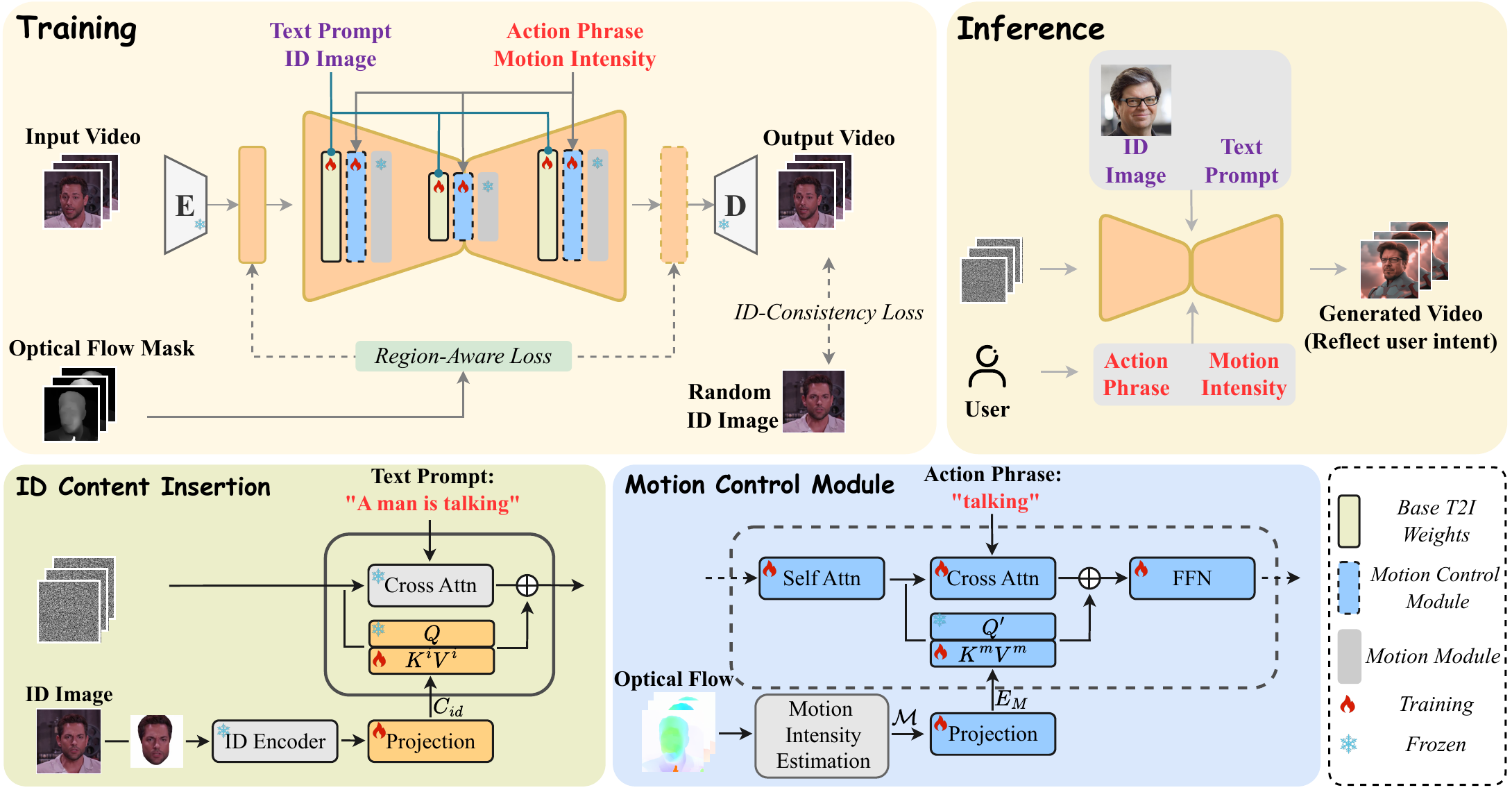}
    \caption{
     Framework overview. Our proposed framework comprises three core components: the ID Content Insertion Module, the Motion Control Module, and a composite loss function. The loss function incorporates a Region-Aware Loss to ensure high motion fidelity and an ID-Consistency Loss to maintain alignment with the reference ID image. During training, motion intensity $\mathcal{M}$ is derived from optical flow. At inference, human animations are generated based on user-defined motion intensity $\mathcal{M}$ and specified action phrases, enabling fine-grained and controllable video synthesis. 
    }
    \label{fig:fig2}
\end{figure*}

\section{Methodology}
\label{methods}

\noindent\textbf{Overall Pipeline.}
Personalized human video generation aims to create vivid clips consistent in character identity and motion based on a reference image and text prompt. To achieve this goal, we propose a framework named \textit{MotionCharacter} which accurately reflects identity information, captures fine-grained motion and maintains smooth visual transitions. Our framework is shown in Fig. \ref{fig:fig2}. Formally, given a reference ID image $\mathcal{I}$, a text prompt $\mathcal{P}$, an action phrase $\mathcal{A}$, and a motion intensity $\mathcal{M}$, the model $\mathcal{F}$ is designed to produce video $\mathcal{V}$ by: 
\begin{equation}
    \mathcal{V} = \mathcal{F}(\mathcal{I}, \mathcal{P}, \mathcal{A}, \mathcal{M}).
    \label{eq:eq1}
\end{equation}

Our methodology is built upon the central principle of disentanglement, designed to establish an orthogonal control space for human video generation. We recognize that to robustly control motion, one must first guarantee the stability of identity, as intense dynamics often corrupt subject-specific features. Therefore, our framework is architected around two synergistic pillars that address this challenge sequentially. First, our ID-Preserving Optimization strategy establishes a stable identity foundation. Second, building upon this foundation, our Motion Control Enhancement module operates to independently manipulate action type and intensity. Furthermore, to enable robust training for such a disentangled framework, we introduce the Human-Motion dataset, specifically curated with dual-track annotations for motion semantics and intensity.

\subsection{ID-Preserving Optimization}
\noindent\textbf{ID Content Insertion. }
Since the adopted pretrained Text-to-Video (T2V) diffusion model~\cite{guo2023animatediff} lacks identity-preserving capabilities, we first intend to introduce an ID Content Insertion Module into the backbone to emphasize identity-specific regions and reduce irrelevant background interference. As illustrated in Fig. \ref{fig:fig2},  the module extracts the identity embedding $C_{id}$ from the reference image $\mathcal{I}$ and injects the identity embedding $C_{id}$ into the diffusion model through cross-attention. 

Specifically, the face region is first isolated from the reference image $\mathcal{I}$  to filter the interference of the background region. Then the face region image is processed in parallel to a pre-trained CLIP image encoder~\cite{radford2021learning} and a face recognition model ArcFace~\cite{deng2019arcface} to obtain the broad contextual identity embeddings $E_{clip}$ and the fine-grained identity embeddings $E_{arc}$, respectively.
To effectively combine global context with fine-grained identity details, we employ cross-attention to fuse the CLIP and ArcFace embeddings:
\begin{equation}
    C_{id} = \text{Proj}(\text{Attn}(E_{arc} W_q', E W_k', E W_v')),
\end{equation}
where $W_q'$, $W_k'$, and $W_v'$ are learnable parameters, with $E_{arc}$ as the query and the combined embedding $E = E_{clip} + E_{arc}$ as the key and value. This fusion allows the detailed ArcFace features to selectively attend to the most relevant contextual information from CLIP embeddings. Following cross-attention, a projection layer $\text{Proj}$ is applied to align the dimension with the text embedding, thereby generating the final identity embedding $C_{id}$ for the reference image  $\mathcal{I}$.

Inspired by recent work on image prompt adapters~\cite{ye2023ip, wang2024instantid}, the identity embedding $C_{id}$ in \textit{MotionCharacter} is regarded as an image prompt embedding and is used alongside text prompt embeddings to provide guidance for the diffusion model. This procedure can be expressed as:
\begin{equation}
    z' = \text{Attn}(Q, K^t, V^t) + \lambda \cdot \text{Attn}(Q, K^i, V^i), 
    \label{eq:adapter_attention_formula}
\end{equation}
where the parameter $\lambda$ controls the balance between text guidance and identity preservation. Here, $Q = z W_q$ is derived from the latent representation $z$, while $K^i = C_{id} W_k^i$ and $V^i = C_{id} W_v^i$ are identity-specific key and value matrices obtained from the identity embedding $C_{id}$ of the reference image $\mathcal{I}$. Similarly, $K^t$, and $V^t$ are the key and value matrices for the text cross-attention.

\noindent\textbf{ID-Consistency Loss.}
To enforce identity preservation at a semantic level, we introduce an ID-Consistency loss that complements the standard pixel-wise MSE objective. The MSE loss, while crucial for visual fidelity, is fundamentally agnostic to high-level concepts like identity. Our proposed loss addresses this by penalizing deviations directly within a learned identity feature space. Specifically, it minimizes the feature-space distance between the reference identity and the generated frames, ensuring that the model maintains the subject's core characteristics even during complex motion.

In practice, at a specific diffusion step $\hat{t}$, the diffusion model can estimate the noise-free latent ${\hat{z}}_0$ from a noisy latent $z_{\hat{t}}$ via the DDIM reversion process. Then, the estimated $\hat{z}_0$ is passed to a VAE decoder to reconstruct the frame, denoted as $X^f$. Therefore, the ID-Consistency loss $\mathcal{L}_{id}$ across the sequence of $N$ frames can be calculated by:
\begin{equation} 
    \mathcal{L}_{id} = 1 - \frac{1}{N} \sum_{i=1}^{N} \frac{\phi(I) \cdot \phi(X^f_i)}{|\phi(I)| |\phi(X^f_i)|},
\end{equation} 
where $\cdot$ denotes the dot product, $\phi$ denotes the pre-trained face recognition backbone~\cite{deng2019arcface}, $\phi(X^f_i)$ and $\phi(I)$ represent the normalized face embedding of each generated frame $i$ and the reference identity image $I$, respectively. $N$ is the total number of frames. 

\subsection{Motion Control Enhancement}
In subject-driven T2V models, achieving precise motion control remains a significant challenge. While previous works~\cite{he2024id} have successfully generated identity-specific videos, they often fail to capture fine-grained motion dynamics, resulting in limited responsiveness to subtle motion cues. 
To address this limitation, we propose a spatial-aware motion control module that explicitly incorporates motion intensity information to enhance controllability. Moreover, a region-aware loss is employed to enhance spatial coherence and realism in dynamic regions such as face. 

We regard the control capacity of the model as lying in two aspects: one is the faithfulness of the motion description, and the other is the magnitude of motion intensity. To achieve this goal, we introduce extra action phrase and motion intensity as the conditions in the proposed model. We first use LMM~\cite{zhu2023minigpt} to automatically generate overall descriptions and action phrases, enriching the dataset with motion-related information and serving as the primary text description $\mathcal{P}$ and action phrase $\mathcal{A}$, as defined in Eq.~\ref{eq:eq1}. Then the action phrase is fed to CLIP text encoder~\cite{radford2021learning} to obtain the action embedding $E_A$ which captures the semantic intent of the motion.

\noindent\textbf{Motion Intensity Estimation.}
The magnitude of motion intensity is challenging to define directly, especially considering the diversity of movements and their varying characteristics. To address this, we employ optical flow estimation to capture motion intensity. Optical flow captures pixel-wise motion between adjacent frames, offering a fine-grained measure of movement that directly reflects the motion of objects in the video. Specifically, given a video clip $\mathcal{V}^{in} = \{v^{in}_i\}_{i=1}^N$, where $N$ is the number of frames, we first extract the optical flow of each pixel between two adjacent frames by:
\begin{equation}
    f_{i,(x,y)} = \Theta(v^{in}_i,v^{in}_{i+1}),
\end{equation}
where $(x,y)$ denotes the position of each pixel, and $\Theta$ is an optical flow estimation model. We use RAFT~\cite{teed2020raft} as $\Theta$ for efficient and accurate optical flow estimation. Then the mean optical flow value $\tau_i$ can be calculated by simply averaging $f_{i,(x,y)}$.
Afterward, we take $\tau_i$ as the threshold to produce binary mask $M_{i,(x,y)}$. Specifically, when the magnitude of the optical flow exceeds $\tau_i$, set the corresponding position in $M_{i,(x,y)}$ to 1; otherwise set it to 0. Consequently, the mean foreground optical flow value $f_{i,fg}$ can be obtained by: 
\begin{equation}
    f_{i,fg} = \frac{1}{S}\sum^H_{x=1}\sum^W_{y=1} f_{i,fg}(x,y) = \frac{1}{S}\sum^H_{x=1}\sum^W_{y=1} M_{i,(x,y)} * f_{i,(x,y)}, 
    \label{eq:foreground_optical_flow}
\end{equation}
where $f_{i,fg}(x,y)$ is the foreground optical flow at each pixel $(x, y)$. $S$ denotes the number of the foreground pixels.

The motion intensity $\mathcal{M}$ of the video is then defined as the average foreground optical flow value across all frames: 
\begin{equation}
   \mathcal{M} = \frac{1}{N-1}\sum^{N-1}_{i=1}f_{i,fg},
\end{equation}
where $\mathcal{M}$ is a scalar representing the motion intensity and $N$ is the number of frames. Subsequently, the motion intensity $\mathcal{M}$ is passed through a motion intensity estimator, utilizing a multi-layer perceptron (MLP) to generate a motion intensity embedding $E_M$ aligned with the dimensionality of the action embedding $E_A$.

\noindent\textbf{Motion Condition Injection.}
As illustrated in Fig. \ref{fig:fig2}, two parallel cross attention modules are adopted in the motion control module to insert the action embedding $E_A$ and motion intensity embedding $E_M$. The process is formally represented as follows: 
 \begin{equation}
   Z'' = \text{Attn}(Q', K^a, V^a) + \alpha \cdot \text{Attn}(Q', K^m, V^m),
\end{equation}
where $Q'=Z'W'_q$ is relevant with the output of ID Content Insertion Module. $K^a$, $V^a$ and $K^m$, $V^m$ are the key-value pairs derived from the action embedding $E_A$ and the motion intensity embedding $E_M$, respectively. 
The parameter $\alpha$ balances the influence of motion intensity within the combined attention output $Z''$ and is set to 1 by default. 

Although both our approach and Stable Video Diffusion (SVD)~\cite{blattmann2023stable} leverage optical flow to characterize motion, a key distinction lies in our explicit decoupling of motion cues. SVD uses motion bucket parameter, which is integrated via additional time embeddings, to coarsely control overall motion magnitude. This results in a global and non-localized representation that may obscure subtle variations (e.g., faces). In contrast, our dual-branch motion condition injection strategy separates semantic guidance (what motion is desired) from quantitative strength (how intense it should be). The use of parallel cross-attention to fuse these decoupled cues is a key mechanism that enables predictable and fine-grained control, allowing our model to capture subtle motion nuances with high fidelity. 

\noindent\textbf{Region-Aware Loss.}
The fluency of the generated video heavily relies on the spatial coherence and realism of dynamic regions, e.g., the face areas. To achieve this goal, we apply a region-aware loss to force the model to focus more on the high-motion regions. Specifically, we normalize the foreground optical flow $f_{i,fg}(x,y)$ defined in Eq. \eqref{eq:foreground_optical_flow} and calculate the optical flow mask $M_{i,\text{norm}}$: 
\begin{equation}
    M_{i,\text{norm}} = \text{clip}\left(\frac{f_{i,fg}(x,y)}{255} + \delta, \, 1.0, \, 1.0 + \delta \right),
\end{equation}
where $\text{clip}(\cdot, \, a, \, b)$ restricts the values to $[a, b]$ and $\delta$ is a scalar offset that modulates the spatial weighting of the optical flow mask. This mask assigns higher weights to regions with significant motion, ensuring that both the primary subject (e.g., faces) and high-dynamic background regions, especially when the text prompt describes a moving background, receive adequate attention. Note that although the optical flow mask assigns non-zero weights to some background regions, it ensures that regions with significant motion from both the subject and background are appropriately emphasized in the loss computation. In contrast to segmentation, depth maps and other similar static scene cues that may overlook such background dynamics, the optical flow mask retains comprehensive motion information across the scene. Then the region-aware loss $\mathcal{L}_{R}$ across all $N$ frames can be compactly defined as: 
\begin{equation}
    \mathcal{L}_{R} = \frac{1}{NH'W'} \sum_{i=1}^{N} \sum_{x=1}^{H'}\sum_{y=1}^{W'} M_{i,\text{norm}} \cdot \left[ \epsilon_i(x,y) - \hat{\epsilon}_i(x,y) \right]^2, 
\end{equation}
where $\epsilon_i(x,y)$ and $\hat{\epsilon}_i(x,y)$ denote the target and predicted noise at location $ (x,y) $, respectively. $H'$ and $W'$ correspond to the resolution of latent.

\subsection{Training Paradigm}\label{sec:training_paradigm}
Our training paradigm is tailored for learning fine-grained, disentangled control. It comprises three integral parts: (1) a dataset with dual-track annotations for motion type and intensity, (2) a mixed image-video strategy for robust generalization and zero-motion calibration, and (3) a loss function that synergistically optimizes for both identity stability and motion fidelity. 

\noindent\textbf{Human-Motion Dataset.}
While foundational, current human video datasets like CelebV-HQ~\cite{zhu2022celebv} and CelebV-Text~\cite{yu2023celebv} lack the fine-grained motion annotations and quality controls needed for training controllable models. Their coarse emotional categories and data artifacts limit the learning of nuanced dynamics. To enable disentangled motion generation, we introduce Human-Motion, comprising 106,292 video clips from various public and private sources. This collection includes VFHQ~\cite{xie2022vfhq} (1,843 clips), CelebV-Text~\cite{yu2023celebv} (52,072 clips), CelebV-HQ~\cite{zhu2022celebv} (31,004 clips), AAHQ~\cite{liu2021blendgan} (17,619 clips), and a private dataset (3,752 clips). Each clip in the Human-Motion dataset was rigorously filtered and re-annotated to ensure high-quality identity and motion information across diverse video formats, resolutions, and styles. 

Human-Motion is distinguished by its dual-track annotation schema, which provides parallel labels for both motion semantics and intensity. Specifically, to enrich the dataset with motion-related information, we used LMM~\cite{zhu2023minigpt} to automatically generate two types of captions for all videos: overall descriptions and action phrases. The overall descriptions provide a general summary of the video's content, while the action phrases offer specific annotations of facial and body movements present in the clips. These captions serve as the primary text description $\mathcal{P}$ and action phrase $\mathcal{A}$ in our framework.

\noindent\textbf{Image-Video Training Strategy.}
To improve the model's generalization across different visual styles, we combined image and video data in training. While realistic videos effectively capture human portraits, they struggle with stylized and artistic content, such as anime. To bridge this gap, we incorporated around 17,619 styled portrait images as static 16-frame videos by replicating each image to simulate a motionless sequence with a motion intensity of 0. It provides a crucial ``zero-intensity calibration" for our motion control module. Training on these static sequences enables the model to learn a robust ``motionless" state representation, which anchors the zero-point of the continuous motion intensity spectrum. This calibration contributes to generating smooth, predictable transitions from complete stillness to dynamic action. Moreover, this approach addresses the challenge of generalizing to stylized portraits by expanding the model's exposure to a wider spectrum of visual characteristics, including variations in texture, color, and artistic exaggeration common in non-realistic styles. By training on both static styled images and dynamic realistic videos, the model learns to preserve identity traits across photorealistic and stylized visual representations, improving its ability to generalize across different visual styles.

\noindent\textbf{Overall Objective.}
The total learning objective combines the Region-Aware Loss, which captures dynamic motion in high-activity regions, and the ID-Consistency Loss, which ensures identity consistency across frames. This dual objective guides the model to preserve both identity and motion fidelity in the generated videos. The total objective function, $\mathcal{L}_{\text{total}}$, is defined as: 
\begin{equation}
    \mathcal{L}_{\text{total}} = \mathcal{L}_{R} + \lambda_{id} \cdot \mathcal{L}_{id}, 
\end{equation}
where $\mathcal{L}_{R}$ and $\mathcal{L}_{id}$ synergistically guide the optimization. $\mathcal{L}_{R}$ ensures motion fidelity by focusing on dynamic regions, while $\mathcal{L}_{id}$ preserves identity consistency across the sequence. The hyperparameter $\lambda_{id}$ mediates the trade-off between these two critical objectives.

\section{Experiments}
\label{sec:experiments}
\subsection{Experiment Setup}
\noindent\textbf{Implementation Details.}
Our implementation is built upon a large-scale pre-trained Video Diffusion Model (VDM)~\cite{guo2023animatediff}.
All experiments were conducted using 8 NVIDIA A100 GPUs (80GB), with the training process taking approximately 24 hours. The batch size was set to 2 for each GPU. The training data consisted of diverse high-quality video clips, which were preprocessed to a resolution of $512\times512$ pixels, with 16 frames sampled per video at a frame rate of 4 frames per second. Data augmentation included random horizontal flipping, resizing, and center cropping to maintain consistent input dimensions. Moreover, text and image dropout rates were set to 0.05, with a 50\% probability of dropping the CLIP embeddings $E_{clip}$. We used the AdamW~\cite{loshchilov2017decoupled} optimizer with a learning rate of $1\times10^{-5}$ and trained the model for 12,000 total steps. For the validation, 16-frame video sequences were generated at a $512\times512$ resolution, applying a standard guidance scale of 8.0 and 30 steps.

\noindent\textbf{Datasets.}
\label{sec:datasets}
For training, we constructed a dataset of 106,292 video clips from various sources. We detail the composition and annotation process in the Training Paradigm subsection. For evaluation, we benchmark our method on the Unsplash-50 test set~\cite{gal2024lcm}, a standard benchmark for ID fidelity containing 50 diverse portraits. To rigorously test performance, we paired each portrait with 140 distinct prompts generated via GPT-4~\cite{achiam2023gpt}, yielding a challenging evaluation suite of 7,000 pairs. 

\noindent\textbf{Evaluation Metrics.}
We assess the quality and consistency of generated videos using six key metrics. The Dover Score~\cite{wu2023dover} assesses overall video quality, considering technical and aesthetic factors. Motion Smoothness~\cite{huang2023vbench} evaluates the continuity of movement between frames, while Dynamic Degree~\cite{huang2023vbench} indicates the extent of motion diversity in the video. CLIP-I~\cite{hessel2021clipscore, xiao2024fastcomposer} measures visual similarity to the reference using the CLIP encoder~\cite{radford2021learning}, and CLIP-T~\cite{hessel2021clipscore, xiao2024fastcomposer} evaluates the alignment between the video content and the text description. To assess identity preservation, we calculate Face Similarity~\cite{xiao2024fastcomposer}, which measures the resemblance between the facial features in the reference image and the generated video.

\subsection{Comparison with Baselines}
We employ four well-known methods in ID-preserving generation task for comparison, i.e., IPA-PlusFace~\cite{ye2023ip}, IPA-FaceID-Portrait~\cite{ye2023ip}, IPA-FaceID-PlusV2~\cite{ye2023ip} and ID-Animator~\cite{he2024id}. They all adopt AnimateDiff~\cite{guo2023animatediff} as the base Text-to-Video generation model. 

\begin{figure}[!t]
    \centering
    \includegraphics[width=\columnwidth]{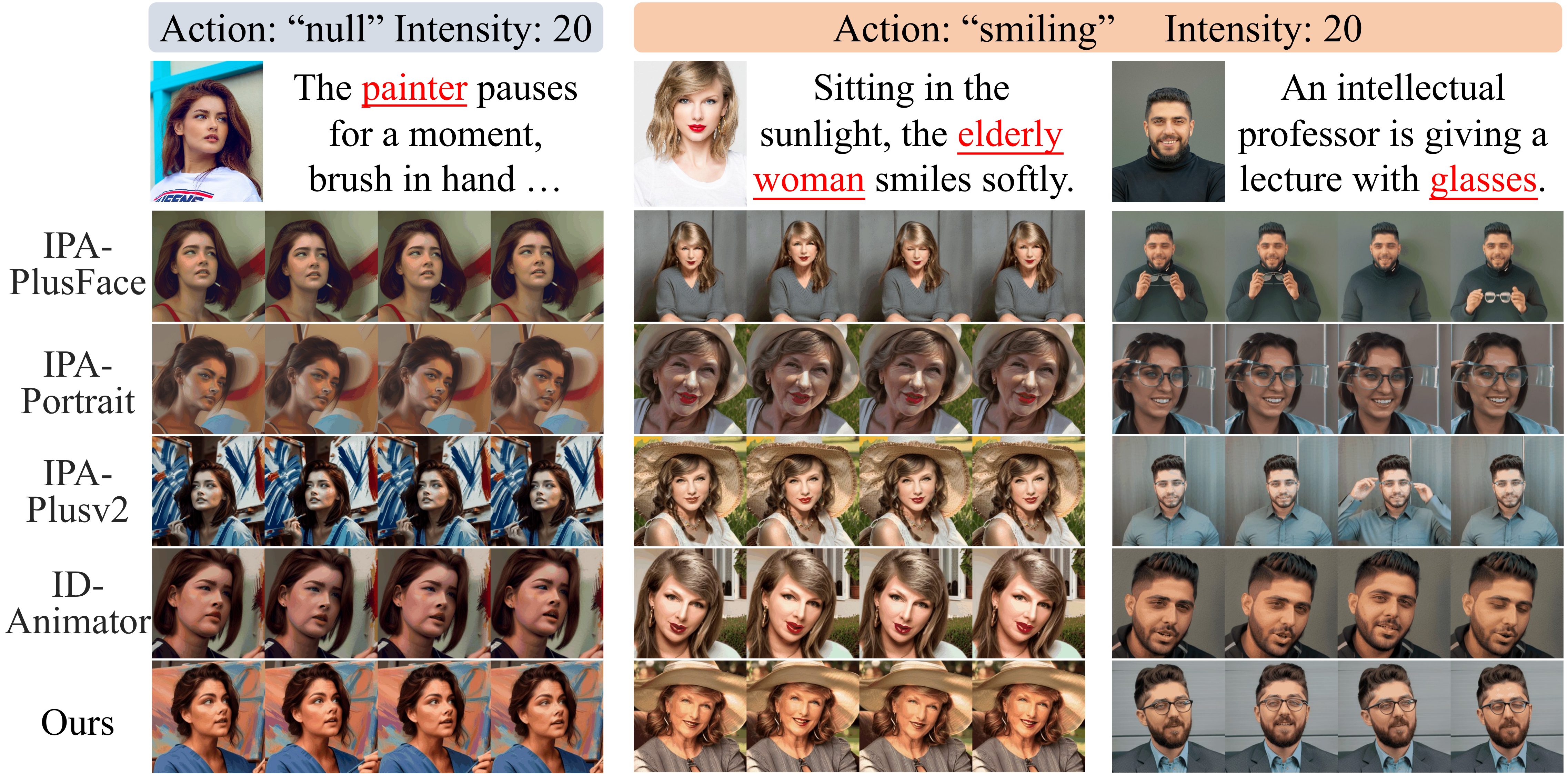}
    \caption{\textbf{Qualitative Comparison.} Comparison of our method with other approaches across diverse prompts and unseen reference images, encompassing various identities (male, female, celebrity, non-celebrity). Each column represents a unique identity and action phrase, with motion intensity fixed at 20 for clarity. ``null" indicates a blank action phrase. Key prompt elements are highlighted in \textcolor{red}{\underline{underline}} to emphasize specific actions or descriptors. For other methods, the action phrase and motion intensity are incorporated with the prompt to guide generation.
    }
    \label{fig:qualitative_comparison}
\end{figure}

\noindent\textbf{Qualitative Comparisons.}
We conduct qualitative comparisons using six diverse identities and prompts generated by GPT-4~\cite{achiam2023gpt}. As shown in Fig.~\ref{fig:qualitative_comparison}, we test a static scenario (null action phrase) to assess baseline ID fidelity and a dynamic scenario (``smiling") to evaluate motion control. The visual results underscore the superiority of \textit{MotionCharacter}. Methods like IPA-FaceID-Portrait exhibit noticeable identity degradation, while others like IPA-PlusFace and IPA-FaceID-PlusV2 suffer from temporal flickering during motion. This indicates their ID representations are not robust enough to withstand dynamic deformations. Furthermore, even the highly competitive ID-Animator fails to bind attributes correctly, omitting ``glasses" from the generated character. This suggests a critical entanglement between its motion and appearance generation modules. In contrast, our method, powered by a dedicated ID-Preserving module and a disentangled motion architecture, successfully maintains identity and all specified attributes while executing the action. 

\begin{table}[!t]
\centering
\scriptsize
\setlength{\tabcolsep}{1.2mm}
\renewcommand{\arraystretch}{0.99}
\begin{tabular}{l!{\vrule width 0.6pt}cccccc}
\Xhline{0.8pt}
\rowcolor{gray!10}
\diagbox[linewidth=0.5pt, linecolor=black, innerleftsep=3pt, innerrightsep=5pt]{\textbf{Method}}{\textbf{Metrics}} & 
\makecell{\textbf{Dover}\\\textbf{Score}$\uparrow$} & 
\makecell{\textbf{Motion}\\\textbf{Smooth.}$\uparrow$} & 
\makecell{\textbf{Dynamic}\\\textbf{Degree}$\uparrow$} & 
\makecell{\textbf{CLIP}\\\textbf{Image}$\uparrow$} & 
\makecell{\textbf{CLIP}\\\textbf{Text}$\uparrow$} & 
\makecell{\textbf{Face}\\\textbf{Sim.}$\uparrow$} \\
\Xhline{0.6pt}
\rowcolor{white} IPA-PlusFace & 0.797 & 0.985 & 0.325 & 0.587 & 0.218 & 0.480 \\
\rowcolor{white} IPA-FaceID-Portrait & 0.849 & 0.984 & 0.191 & 0.545 & \underline{0.223} & 0.531 \\
\rowcolor{white} IPA-FaceID-PlusV2 & 0.813 & \underline{0.987} & 0.085 & 0.575 & 0.217 & \textbf{0.617} \\
\rowcolor{white} ID-Animator & \underline{0.857} & 0.979 & \underline{0.433} & \underline{0.607} & 0.204 & 0.546 \\
\rowcolor{white} \textbf{Ours} & \textbf{0.869} & \textbf{0.998} & \textbf{0.449} & \textbf{0.633} & \textbf{0.227} & \underline{0.609} \\
\Xhline{0.8pt}
\end{tabular}
\caption{Quantitative comparison with state-of-the-art methods. Higher values ($\uparrow$) indicate better performance. \textbf{Bold} and \underline{underlined} numbers denote the best and second-best results, respectively. All methods use an empty action phrase with motion intensity set to 20 for fair comparison.}
\label{tab:comparison}
\end{table}

\noindent\textbf{Quantitative Comparisons.}
The quantitative results in Table~\ref{tab:comparison} further substantiate our findings and reveal a critical challenge in existing methods: the inherent trade-off between identity preservation and motion dynamics. Specifically, IPA-FaceID-PlusV2 achieves the highest Face Similarity (0.617) but at the cost of producing nearly static videos, evidenced by its extremely low Dynamic Degree (0.085). Conversely, ID-Animator attains a high Dynamic Degree (0.433) but with significantly compromised identity fidelity and weaker text alignment (CLIP-T of 0.204), which confirms our qualitative observations. This demonstrates that current methods are forced to sacrifice either motion or identity. In contrast, \textit{MotionCharacter} breaks this trade-off. Our method achieves state-of-the-art scores in five out of six metrics, including overall quality (Dover), motion attributes, and content consistency (CLIP-I, CLIP-T). Crucially, we attain a Face Similarity score (0.609) that is highly competitive with the top score, while simultaneously achieving the highest Dynamic Degree (0.449). This result is a powerful validation of our core hypothesis: by explicitly disentangling motion control from identity representation, our framework can generate highly dynamic and expressive videos without compromising identity fidelity. The negligible 1.3\% difference in Face Similarity is a small price for a staggering 428\% improvement in Dynamic Degree compared to the leading ID-preserving method.

\begin{figure}[t!]
    \centering
    \includegraphics[width=\linewidth]{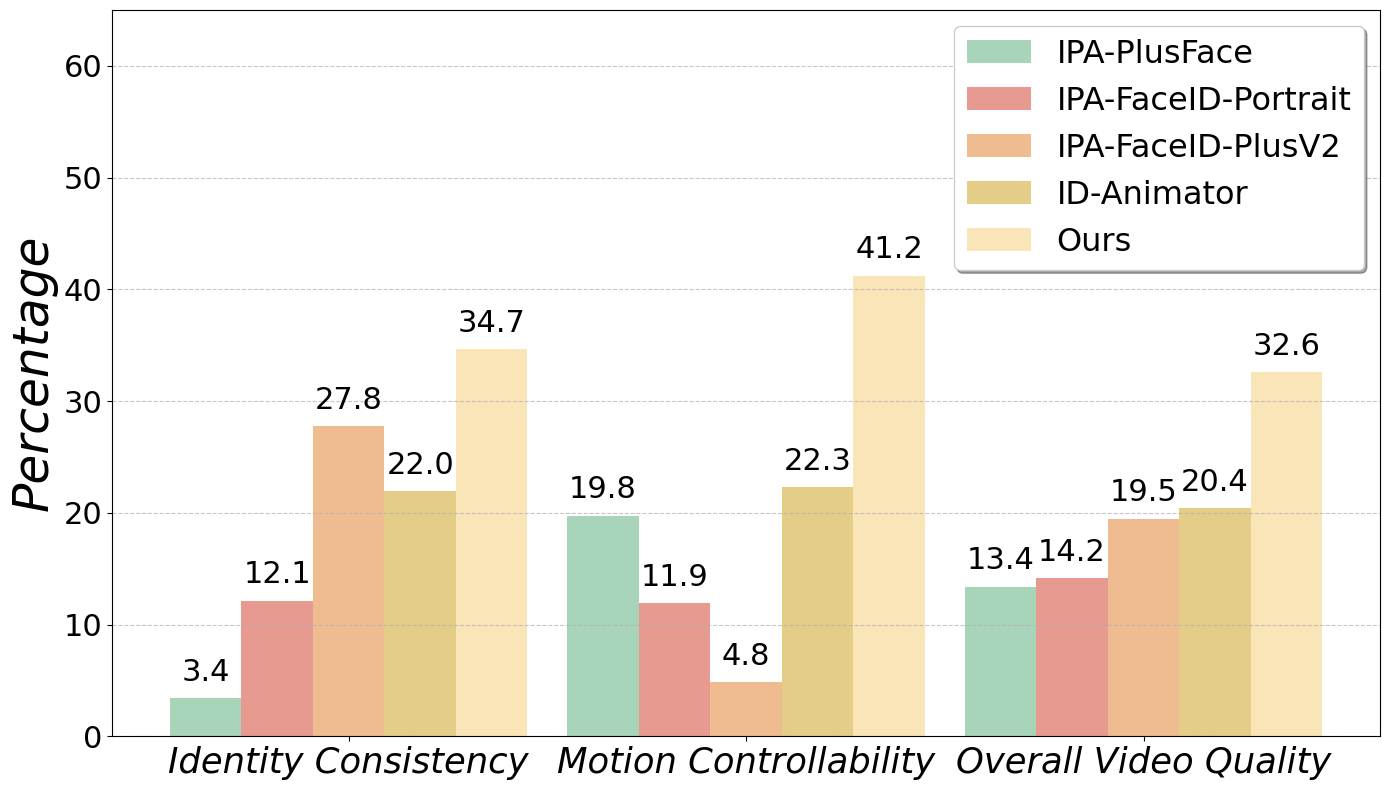}
    \caption{User study results comparing our method with baselines across three evaluation criteria: identity consistency, motion controllability, and overall video quality. }
    \label{fig:user_study}
\end{figure}

\noindent\textbf{User Study.}
Recognizing that CLIP scores~\cite{hessel2021clipscore, xiao2024fastcomposer} may not fully align with human perception~\cite{molad2023dreamix, wang2023zero}, we conducted a user study to validate our quantitative findings. The study involved 10 expert raters from enterprises and academic labs, each with expertise in generative models, evaluating 100 videos across three key dimensions: Identity Consistency, Motion Controllability, and Overall Video Quality. For each video, participants selected the best performing method among all approaches, with videos presented in randomized order to prevent bias. This resulted in 3,000 total ratings (10 raters $\times$ 100 videos $\times$ 3 criteria). As shown in Fig.~\ref{fig:user_study}, our method consistently received the highest preference across all evaluation criteria. Notably, the slight discrepancy between Face Similarity metrics and human perception of identity preservation can be attributed to varying levels of motion richness. While methods like ID-Animator and IPA achieve high similarity scores through strict identity constraints that limit facial dynamics, our approach generates more varied and expressive motions while maintaining perceptually consistent identity. This human evaluation validates our quantitative results, confirming that our decoupled control framework achieves a superior balance between identity preservation and motion expressiveness.

\subsection{Ablation Study}
\noindent\textbf{Region-Aware Loss.}
As shown in Table~\ref{tab:ablation_region}, integrating only the Region-Aware Loss ($\mathcal{L}_{R}$) yields substantial improvements in motion-related metrics compared to the vanilla baseline. Specifically, it boosts the Dover Score by a significant +0.059 and increases the Dynamic Degree by +0.064, while also enhancing Motion Smoothness. This quantitatively confirms that by forcing the model to focus its objective on high-motion areas, $\mathcal{L}_{R}$ is highly effective at improving the clarity and quality of dynamic movements. Notably, its impact on Face Similarity is minimal, underscoring its specific role in refining motion rather than identity.

\begin{table}[!t]
\centering
\scriptsize
\setlength{\tabcolsep}{1.0mm}
\renewcommand{\arraystretch}{0.99}
\begin{tabular}{cc!{\vrule width 0.6pt}cccccc}
\Xhline{0.8pt}
\rowcolor{gray!4}
\multicolumn{2}{c!{\vrule width 0.6pt}}{\textbf{Loss}} & \multicolumn{6}{c}{\textbf{Metrics}} \\
\Xcline{1-2}{0.6pt} \Xcline{3-8}{0.6pt}
\rowcolor{gray!10} \(\mathcal{L}_{R}\) & \(\mathcal{L}_{id}\) & 
\makecell{\textbf{Dover}\\\textbf{Score}$\uparrow$} & 
\makecell{\textbf{Motion}\\\textbf{Smoothness}$\uparrow$} & 
\makecell{\textbf{Dynamic}\\\textbf{Degree}$\uparrow$} & 
\makecell{\textbf{CLIP}\\\textbf{Image}$\uparrow$} & 
\makecell{\textbf{CLIP}\\\textbf{Text}$\uparrow$} & 
\makecell{\textbf{Face}\\\textbf{Similarity}$\uparrow$} \\
\Xhline{0.6pt}
\rowcolor{white} \xmark & \xmark & 0.801 & 0.978 & 0.355 & 0.599 & 0.219 & 0.484 \\
\rowcolor{white} \xmark & \cmark & 0.810 & 0.983 & 0.359 & 0.627 & 0.220 & 0.588 \\
\rowcolor{white} \cmark & \xmark & 0.860 & 0.995 & 0.419 & 0.611 & 0.224 & 0.500 \\
\rowcolor{white} \cmark & \cmark & \textbf{0.869} & \textbf{0.998} & \textbf{0.449} & \textbf{0.633} & \textbf{0.227} & \textbf{0.609} \\
\Xhline{0.8pt}
\end{tabular}
\caption{Ablation study of the Region-Aware Loss $\mathcal{L}_{R}$ and the ID-Consistency Loss $\mathcal{L}_{id}$.}
\label{tab:ablation_region}
\end{table}

\begin{table}[t]
\centering
\scriptsize
\setlength{\tabcolsep}{1.2mm}
\renewcommand{\arraystretch}{0.99}
\begin{tabular}{c!{\vrule width 0.6pt}cccccc}
\Xhline{0.8pt}
\rowcolor{gray!4}
\multicolumn{1}{c!{\vrule width 0.6pt}}{\textbf{Module}} & \multicolumn{6}{c}{\textbf{Metrics}} \\
\Xcline{1-1}{0.6pt} \Xcline{2-7}{0.6pt}
\rowcolor{gray!10} \textbf{MCM} & 
\makecell{\textbf{Dover}\\\textbf{Score}$\uparrow$} & 
\makecell{\textbf{Motion}\\\textbf{Smoothness}$\uparrow$} & 
\makecell{\textbf{Dynamic}\\\textbf{Degree}$\uparrow$} & 
\makecell{\textbf{CLIP}\\\textbf{Image}$\uparrow$} & 
\makecell{\textbf{CLIP}\\\textbf{Text}$\uparrow$} & 
\makecell{\textbf{Face}\\\textbf{Similarity}$\uparrow$} \\
\Xhline{0.6pt}
\rowcolor{white} \xmark & 0.805 & 0.978 & 0.245 & 0.563 & 0.204 & 0.601 \\
\rowcolor{white} \cmark & \textbf{0.869} & \textbf{0.998} & \textbf{0.449} & \textbf{0.633} & \textbf{0.227} & \textbf{0.609} \\
\Xhline{0.8pt}
\end{tabular}
\caption{Ablation study of Motion Control Module (MCM).}
\label{tab:ablation_module}
\end{table}
\noindent\textbf{ID-Consistency Loss.}
The ID-Consistency Loss ($\mathcal{L}_{id}$) is designed to anchor the subject's identity. When added alone, it produces a massive gain of +0.104 in Face Similarity, proving its efficacy in robust identity preservation. However, the most compelling finding is the synergistic effect when both losses are combined. The full model not only achieves the best scores across all metrics but also surpasses the performance of individual components in their respective domains. For instance, the final Face Similarity (0.609) is higher than with $\mathcal{L}_{id}$ alone (0.588), and the Dynamic Degree (0.449) is greater than with $\mathcal{L}_{R}$ alone (0.419). This powerful synergy validates our core hypothesis: a stable identity foundation provided by $\mathcal{L}_{id}$ enables $\mathcal{L}_{R}$ to sculpt more expressive and high-fidelity motion, demonstrating that both losses are critical and complementary for achieving state-of-the-art results.

\noindent\textbf{Motion Control Module.} 
Table~\ref{tab:ablation_module} demonstrates the substantial impact of our Motion Control Module (MCM). Its introduction leads to consistent improvements across all metrics, with particularly notable gains in Dynamic Degree (83.3\% increase from 0.245 to 0.449). This dramatic improvement in motion expressiveness, achieved while maintaining face similarity (0.609), validates our module's ability to enhance motion control without compromising identity preservation.

\section{Conclusions}
In this paper, we presented \textit{MotionCharacter}, a high-fidelity human video generation framework that successfully addresses the fundamental challenge of achieving fine-grained motion control while ensuring strict identity preservation. The key to our approach is a disentangled architecture where two parallel mechanisms operate. The first mechanism anchors the subject's appearance using a dedicated ID Content Insertion Module supervised by an ID-Consistency Loss. The second enables precise control over action and intensity via the Motion Control Module, with a Region-Aware Loss ensuring clarity in dynamic areas. This entire framework is empowered by our Human-Motion dataset, whose fine-grained annotations are crucial for learning such disentangled representations. Exhaustive experiments validate that \textit{MotionCharacter} not only outperforms existing methods in overall quality but also effectively breaks the critical trade-off between motion dynamics and identity fidelity. 

\section{Acknowledgments}
This work was supported by the National Natural Science Foundation of China (Grant U25A20403), the Natural Science Foundation of Hubei Province of China (No. 2024AFB545). We also thank Meituan for their support. 
\bibliography{aaai2026}

@article{loshchilov2017decoupled,
  title={Decoupled weight decay regularization},
  author={Loshchilov, I},
  journal={arXiv preprint arXiv:1711.05101},
  year={2017}
}

@inproceedings{deng2019arcface,
  title={Arcface: Additive angular margin loss for deep face recognition},
  author={Deng, Jiankang and Guo, Jia and Xue, Niannan and Zafeiriou, Stefanos},
  booktitle={Proceedings of the IEEE/CVF conference on computer vision and pattern recognition},
  pages={4690--4699},
  year={2019}
}

@inproceedings{teed2020raft,
  title={Raft: Recurrent all-pairs field transforms for optical flow},
  author={Teed, Zachary and Deng, Jia},
  booktitle={Computer Vision--ECCV 2020: 16th European Conference, Glasgow, UK, August 23--28, 2020, Proceedings, Part II 16},
  pages={402--419},
  year={2020},
  organization={Springer}
}

@article{ho2020denoising,
  title={Denoising diffusion probabilistic models},
  author={Ho, Jonathan and Jain, Ajay and Abbeel, Pieter},
  journal={Advances in neural information processing systems},
  volume={33},
  pages={6840--6851},
  year={2020}
}

@inproceedings{
song2020score,
title={Score-Based Generative Modeling through Stochastic Differential Equations},
author={Yang Song and Jascha Sohl-Dickstein and Diederik P Kingma and Abhishek Kumar and Stefano Ermon and Ben Poole},
booktitle={International Conference on Learning Representations},
year={2021},
url={https://openreview.net/forum?id=PxTIG12RRHS}
}

@article{liu2021blendgan,
  title={Blendgan: Implicitly gan blending for arbitrary stylized face generation},
  author={Liu, Mingcong and Li, Qiang and Qin, Zekui and Zhang, Guoxin and Wan, Pengfei and Zheng, Wen},
  journal={Advances in Neural Information Processing Systems},
  volume={34},
  pages={29710--29722},
  year={2021}
}

@inproceedings{hessel2021clipscore,
  title={CLIPScore: A Reference-free Evaluation Metric for Image Captioning},
  author={Hessel, Jack and Holtzman, Ari and Forbes, Maxwell and Le Bras, Ronan and Choi, Yejin},
  booktitle={Proceedings of the 2021 Conference on Empirical Methods in Natural Language Processing},
  pages={7514--7528},
  year={2021}
}

@article{ho2022video,
  title={Video diffusion models},
  author={Ho, Jonathan and Salimans, Tim and Gritsenko, Alexey and Chan, William and Norouzi, Mohammad and Fleet, David J},
  journal={Advances in Neural Information Processing Systems},
  volume={35},
  pages={8633--8646},
  year={2022}
}

@article{ho2022imagen,
  title={Imagen video: High definition video generation with diffusion models},
  author={Ho, Jonathan and Chan, William and Saharia, Chitwan and Whang, Jay and Gao, Ruiqi and Gritsenko, Alexey and Kingma, Diederik P and Poole, Ben and Norouzi, Mohammad and Fleet, David J and others},
  journal={arXiv preprint arXiv:2210.02303},
  year={2022}
}

@article{singer2022make,
  title={Make-a-video: Text-to-video generation without text-video data},
  author={Singer, Uriel and Polyak, Adam and Hayes, Thomas and Yin, Xi and An, Jie and Zhang, Songyang and Hu, Qiyuan and Yang, Harry and Ashual, Oron and Gafni, Oran and others},
  journal={arXiv preprint arXiv:2209.14792},
  year={2022}
}

@Misc{xFormers2022,
  author =       {Benjamin Lefaudeux and Francisco Massa and Diana Liskovich and Wenhan Xiong and Vittorio Caggiano and Sean Naren and Min Xu and Jieru Hu and Marta Tintore and Susan Zhang and Patrick Labatut and Daniel Haziza and Luca Wehrstedt and Jeremy Reizenstein and Grigory Sizov},
  title =        {xFormers: A modular and hackable Transformer modelling library},
  howpublished = {\url{https://github.com/facebookresearch/xformers}},
  year =         {2022}
}

@article{zhou2022magicvideo,
  title={Magicvideo: Efficient video generation with latent diffusion models},
  author={Zhou, Daquan and Wang, Weimin and Yan, Hanshu and Lv, Weiwei and Zhu, Yizhe and Feng, Jiashi},
  journal={arXiv preprint arXiv:2211.11018},
  year={2022}
}

@inproceedings{xie2022vfhq,
  title={Vfhq: A high-quality dataset and benchmark for video face super-resolution},
  author={Xie, Liangbin and Wang, Xintao and Zhang, Honglun and Dong, Chao and Shan, Ying},
  booktitle={Proceedings of the IEEE/CVF Conference on Computer Vision and Pattern Recognition},
  pages={657--666},
  year={2022}
}

@inproceedings{zhu2022celebv,
  title={CelebV-HQ: A large-scale video facial attributes dataset},
  author={Zhu, Hao and Wu, Wayne and Zhu, Wentao and Jiang, Liming and Tang, Siwei and Zhang, Li and Liu, Ziwei and Loy, Chen Change},
  booktitle={European conference on computer vision},
  pages={650--667},
  year={2022},
  organization={Springer}
}

@inproceedings{rombach2022high,
  title={High-resolution image synthesis with latent diffusion models},
  author={Rombach, Robin and Blattmann, Andreas and Lorenz, Dominik and Esser, Patrick and Ommer, Bj{\"o}rn},
  booktitle={Proceedings of the IEEE/CVF conference on computer vision and pattern recognition},
  pages={10684--10695},
  year={2022}
}

@inproceedings{yu2023celebv,
  title={Celebv-text: A large-scale facial text-video dataset},
  author={Yu, Jianhui and Zhu, Hao and Jiang, Liming and Loy, Chen Change and Cai, Weidong and Wu, Wayne},
  booktitle={Proceedings of the IEEE/CVF Conference on Computer Vision and Pattern Recognition},
  pages={14805--14814},
  year={2023}
}

@inproceedings{
guo2023animatediff,
title={AnimateDiff: Animate Your Personalized Text-to-Image Diffusion Models without Specific Tuning},
author={Yuwei Guo and Ceyuan Yang and Anyi Rao and Zhengyang Liang and Yaohui Wang and Yu Qiao and Maneesh Agrawala and Dahua Lin and Bo Dai},
booktitle={International Conference on Learning Representations},
year={2024},
url={https://openreview.net/forum?id=Fx2SbBgcte}
}

@inproceedings{blattmann2023align,
  title={Align your latents: High-resolution video synthesis with latent diffusion models},
  author={Blattmann, Andreas and Rombach, Robin and Ling, Huan and Dockhorn, Tim and Kim, Seung Wook and Fidler, Sanja and Kreis, Karsten},
  booktitle={Proceedings of the IEEE/CVF Conference on Computer Vision and Pattern Recognition},
  pages={22563--22575},
  year={2023}
}

@article{blattmann2023stable,
  title={Stable video diffusion: Scaling latent video diffusion models to large datasets},
  author={Blattmann, Andreas and Dockhorn, Tim and Kulal, Sumith and Mendelevitch, Daniel and Kilian, Maciej and Lorenz, Dominik and Levi, Yam and English, Zion and Voleti, Vikram and Letts, Adam and others},
  journal={arXiv preprint arXiv:2311.15127},
  year={2023}
}

@article{wang2023modelscope,
  title={Modelscope text-to-video technical report},
  author={Wang, Jiuniu and Yuan, Hangjie and Chen, Dayou and Zhang, Yingya and Wang, Xiang and Zhang, Shiwei},
  journal={arXiv preprint arXiv:2308.06571},
  year={2023}
}

@article{ye2023ip,
  title={Ip-adapter: Text compatible image prompt adapter for text-to-image diffusion models},
  author={Ye, Hu and Zhang, Jun and Liu, Sibo and Han, Xiao and Yang, Wei},
  journal={arXiv preprint arXiv:2308.06721},
  year={2023}
}

@article{chen2023videocrafter1,
  title={Videocrafter1: Open diffusion models for high-quality video generation},
  author={Chen, Haoxin and Xia, Menghan and He, Yingqing and Zhang, Yong and Cun, Xiaodong and Yang, Shaoshu and Xing, Jinbo and Liu, Yaofang and Chen, Qifeng and Wang, Xintao and others},
  journal={arXiv preprint arXiv:2310.19512},
  year={2023}
}

@inproceedings{ruiz2023dreambooth,
  title={Dreambooth: Fine tuning text-to-image diffusion models for subject-driven generation},
  author={Ruiz, Nataniel and Li, Yuanzhen and Jampani, Varun and Pritch, Yael and Rubinstein, Michael and Aberman, Kfir},
  booktitle={Proceedings of the IEEE/CVF conference on computer vision and pattern recognition},
  pages={22500--22510},
  year={2023}
}

@article{achiam2023gpt,
  title={Gpt-4 technical report},
  author={Achiam, Josh and Adler, Steven and Agarwal, Sandhini and Ahmad, Lama and Akkaya, Ilge and Aleman, Florencia Leoni and Almeida, Diogo and Altenschmidt, Janko and Altman, Sam and Anadkat, Shyamal and others},
  journal={arXiv preprint arXiv:2303.08774},
  year={2023}
}

@inproceedings{huang2023vbench,
  title={Vbench: Comprehensive benchmark suite for video generative models},
  author={Huang, Ziqi and He, Yinan and Yu, Jiashuo and Zhang, Fan and Si, Chenyang and Jiang, Yuming and Zhang, Yuanhan and Wu, Tianxing and Jin, Qingyang and Chanpaisit, Nattapol and others},
  booktitle={Proceedings of the IEEE/CVF Conference on Computer Vision and Pattern Recognition},
  pages={21807--21818},
  year={2024}
}

@inproceedings{wu2023dover,
  title={Exploring video quality assessment on user generated contents from aesthetic and technical perspectives},
  author={Wu, Haoning and Zhang, Erli and Liao, Liang and Chen, Chaofeng and Hou, Jingwen and Wang, Annan and Sun, Wenxiu and Yan, Qiong and Lin, Weisi},
  booktitle={Proceedings of the IEEE/CVF International Conference on Computer Vision},
  pages={20144--20154},
  year={2023}
}

@article{zhu2023minigpt,
  title={Minigpt-4: Enhancing vision-language understanding with advanced large language models},
  author={Zhu, Deyao and Chen, Jun and Shen, Xiaoqian and Li, Xiang and Elhoseiny, Mohamed},
  journal={arXiv preprint arXiv:2304.10592},
  year={2023}
}

@article{molad2023dreamix,
  title={Dreamix: Video diffusion models are general video editors},
  author={Molad, Eyal and Horwitz, Eliahu and Valevski, Dani and Acha, Alex Rav and Matias, Yossi and Pritch, Yael and Leviathan, Yaniv and Hoshen, Yedid},
  journal={arXiv preprint arXiv:2302.01329},
  year={2023}
}

@inproceedings{wang2023exploring,
  title={Exploring clip for assessing the look and feel of images},
  author={Wang, Jianyi and Chan, Kelvin CK and Loy, Chen Change},
  booktitle={Proceedings of the AAAI conference on artificial intelligence},
  volume={37},
  number={2},
  pages={2555--2563},
  year={2023}
}

@inproceedings{gal2024lcm,
  title={Lcm-lookahead for encoder-based text-to-image personalization},
  author={Gal, Rinon and Lichter, Or and Richardson, Elad and Patashnik, Or and Bermano, Amit H and Chechik, Gal and Cohen-Or, Daniel},
  booktitle={European Conference on Computer Vision},
  pages={322--340},
  year={2024},
  organization={Springer}
}

@article{wang2023zero,
  title={Zero-shot video editing using off-the-shelf image diffusion models},
  author={Wang, Wen and Jiang, Yan and Xie, Kangyang and Liu, Zide and Chen, Hao and Cao, Yue and Wang, Xinlong and Shen, Chunhua},
  journal={arXiv preprint arXiv:2303.17599},
  year={2023}
}

@inproceedings{chen2024videocrafter2,
  title={Videocrafter2: Overcoming data limitations for high-quality video diffusion models},
  author={Chen, Haoxin and Zhang, Yong and Cun, Xiaodong and Xia, Menghan and Wang, Xintao and Weng, Chao and Shan, Ying},
  booktitle={Proceedings of the IEEE/CVF Conference on Computer Vision and Pattern Recognition},
  pages={7310--7320},
  year={2024}
}

@inproceedings{li2024photomaker,
  title={Photomaker: Customizing realistic human photos via stacked id embedding},
  author={Li, Zhen and Cao, Mingdeng and Wang, Xintao and Qi, Zhongang and Cheng, Ming-Ming and Shan, Ying},
  booktitle={Proceedings of the IEEE/CVF Conference on Computer Vision and Pattern Recognition},
  pages={8640--8650},
  year={2024}
}

@article{wang2024instantid,
  title={Instantid: Zero-shot identity-preserving generation in seconds},
  author={Wang, Qixun and Bai, Xu and Wang, Haofan and Qin, Zekui and Chen, Anthony and Li, Huaxia and Tang, Xu and Hu, Yao},
  journal={arXiv preprint arXiv:2401.07519},
  year={2024}
}

@article{xiao2024fastcomposer,
  title={Fastcomposer: Tuning-free multi-subject image generation with localized attention},
  author={Xiao, Guangxuan and Yin, Tianwei and Freeman, William T and Durand, Fr{\'e}do and Han, Song},
  journal={International Journal of Computer Vision},
  pages={1--20},
  year={2024},
  publisher={Springer}
}

@inproceedings{jiang2024videobooth,
  title={Videobooth: Diffusion-based video generation with image prompts},
  author={Jiang, Yuming and Wu, Tianxing and Yang, Shuai and Si, Chenyang and Lin, Dahua and Qiao, Yu and Loy, Chen Change and Liu, Ziwei},
  booktitle={Proceedings of the IEEE/CVF Conference on Computer Vision and Pattern Recognition},
  pages={6689--6700},
  year={2024}
}

@inproceedings{wei2024dreamvideo,
  title={Dreamvideo: Composing your dream videos with customized subject and motion},
  author={Wei, Yujie and Zhang, Shiwei and Qing, Zhiwu and Yuan, Hangjie and Liu, Zhiheng and Liu, Yu and Zhang, Yingya and Zhou, Jingren and Shan, Hongming},
  booktitle={Proceedings of the IEEE/CVF Conference on Computer Vision and Pattern Recognition},
  pages={6537--6549},
  year={2024}
}

@article{ma2024magic,
  title={Magic-me: Identity-specific video customized diffusion},
  author={Ma, Ze and Zhou, Daquan and Yeh, Chun-Hsiao and Wang, Xue-She and Li, Xiuyu and Yang, Huanrui and Dong, Zhen and Keutzer, Kurt and Feng, Jiashi},
  journal={arXiv preprint arXiv:2402.09368},
  year={2024}
}

@article{wu2024customcrafter,
  title={Customcrafter: Customized video generation with preserving motion and concept composition abilities},
  author={Wu, Tao and Zhang, Yong and Wang, Xintao and Zhou, Xianpan and Zheng, Guangcong and Qi, Zhongang and Shan, Ying and Li, Xi},
  journal={arXiv preprint arXiv:2408.13239},
  year={2024}
}

@article{he2024id,
  title={ID-Animator: Zero-Shot Identity-Preserving Human Video Generation},
  author={He, Xuanhua and Liu, Quande and Qian, Shengju and Wang, Xin and Hu, Tao and Cao, Ke and Yan, Keyu and Zhou, Man and Zhang, Jie},
  journal={arXiv preprint arXiv:2404.15275},
  year={2024}
}

@misc{JaidedAI_EasyOCR,
  author = {JaidedAI},
  title = {EasyOCR},
  howpublished = {\url{https://github.com/JaidedAI/EasyOCR}},
  year = {2024}
}

@article{wang2023lavie,
  title={Lavie: High-quality video generation with cascaded latent diffusion models},
  author={Wang, Yaohui and Chen, Xinyuan and Ma, Xin and Zhou, Shangchen and Huang, Ziqi and Wang, Yi and Yang, Ceyuan and He, Yinan and Yu, Jiashuo and Yang, Peiqing and others},
  journal={International Journal of Computer Vision},
  pages={1--20},
  year={2024},
  publisher={Springer}
}

@inproceedings{radford2021learning,
  title={Learning transferable visual models from natural language supervision},
  author={Radford, Alec and Kim, Jong Wook and Hallacy, Chris and Ramesh, Aditya and Goh, Gabriel and Agarwal, Sandhini and Sastry, Girish and Askell, Amanda and Mishkin, Pamela and Clark, Jack and others},
  booktitle={International conference on machine learning},
  pages={8748--8763},
  year={2021},
}

@article{wan2025wan,
  title={Wan: Open and advanced large-scale video generative models},
  author={Wan, Team and Wang, Ang and Ai, Baole and Wen, Bin and Mao, Chaojie and Xie, Chen-Wei and Chen, Di and Yu, Feiwu and Zhao, Haiming and Yang, Jianxiao and others},
  journal={arXiv preprint arXiv:2503.20314},
  year={2025}
}

\clearpage
\setcounter{page}{1}

\setcounter{section}{0}
\setcounter{table}{0}
\setcounter{figure}{0}
\renewcommand{\thefigure}{\Roman{figure}}
\renewcommand{\thesection}{\Alph{section}}
\renewcommand{\thetable}{\Roman{table}}

\section{Additional Dataset Analysis}
\label{sec:data}
\begin{figure}[!h]
    \centering
    \includegraphics[width=0.5\columnwidth]{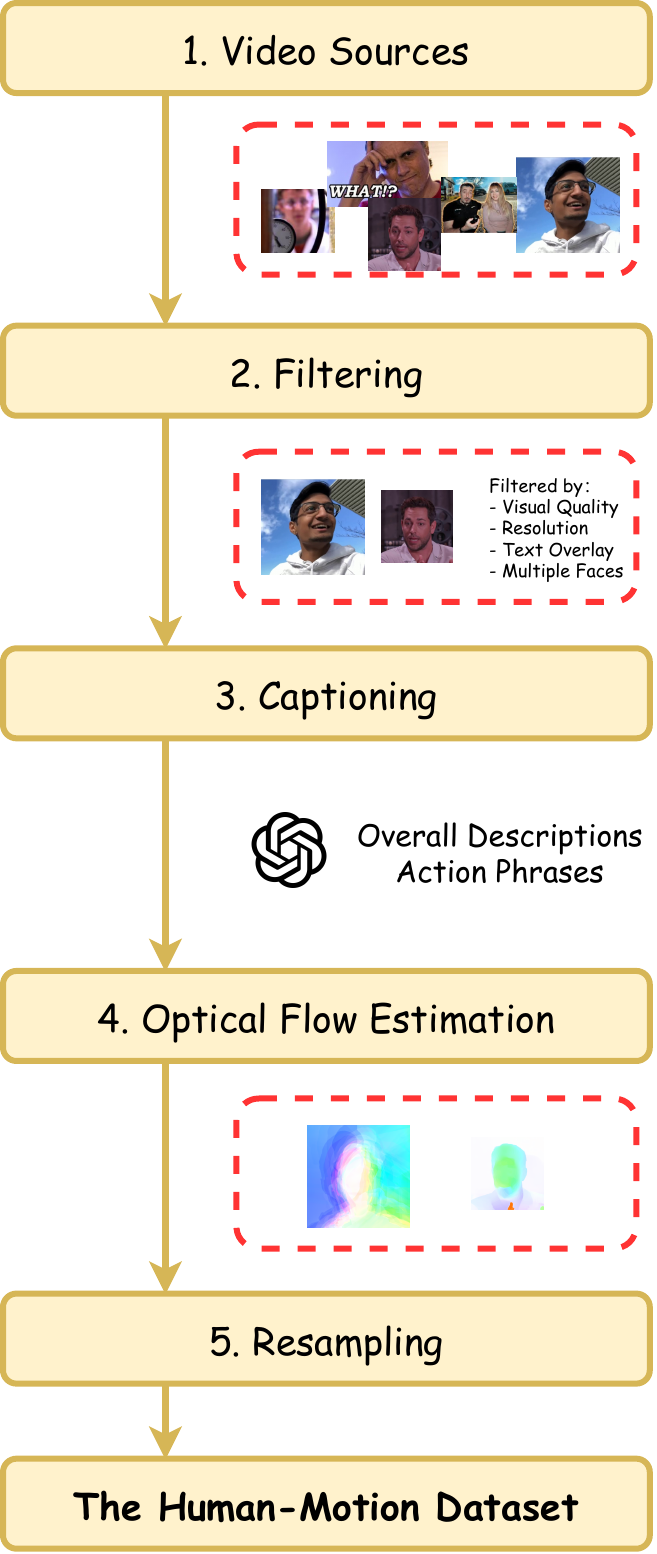}
    \caption{The building process of the Human-Motion Dataset.}
    \label{fig:dataset_process}
\end{figure}

To build the Human-Motion dataset, a multi-step pipeline was developed to ensure the collection of high-quality video clips. The detailed process of building the dataset is illustrated in Fig. \ref{fig:dataset_process}.

\subsection{Video Sources} Our data sources comprise video clips from diverse origins, including VFHQ~\cite{xie2022vfhq}, CelebV-Text~\cite{yu2023celebv}, CelebV-HQ~\cite{zhu2022celebv}, AAHQ~\cite{liu2021blendgan}, and a private dataset, Sing Videos. Each clip was carefully filtered and re-annotated to ensure high-quality identity and motion information across various formats, resolutions, and styles.

\subsection{Filtering Process} To maintain data quality, a multi-step filtering process was applied:
\begin{itemize}
    \item \textbf{Visual Quality Check}: We used CLIP Image Quality Assessment~\cite{wang2023exploring} (CLIP-IQA) to evaluate visual quality by sampling one frame per clip, discarding videos with frames of low quality.
    \item \textbf{Resolution Filter}: Videos with resolutions below 512 pixels were removed to uphold visual standards.
    \item \textbf{Text Overlay Detection}: EasyOCR~\cite{JaidedAI_EasyOCR} was used to detect excessive subtitles or text overlays, filtering out obstructed frames.
    \item \textbf{Face Detection}: Videos containing multiple faces or low face detection confidence were discarded to ensure each video contains a single, clearly detectable person.
\end{itemize}

\subsection{Captioning} To enrich motion-related data, we utilized MiniGPT~\cite{zhu2023minigpt} to automatically generate two types of captions for each video:
\begin{itemize}
    \item \textbf{Overall Descriptions $\mathcal{P}$}: General summaries of the video content.
    \item \textbf{Action Phrases $\mathcal{A}$}: Detailed annotations of facial and body movements, serving as action phrases $\mathcal{A}$ in our framework.
\end{itemize}
This dual-captioning strategy enhances the dataset by providing both global context and specific motion dynamics, equipping the model to generate identity-consistent human video clips with controllable action phrases.

\subsection{Optical Flow Estimation}
Optical flow estimation for video is performed using the RAFT~\cite{teed2020raft} model on consecutive frames to compute motion information. The RAFT model calculates the optical flow field, representing pixel displacements between frames. In this study, we use the extracted optical flow to obtain motion information for each video segment, enabling accurate motion modeling and control. These optical flows are used to compute motion intensity during the training phase and are also utilized in optimizing the loss function.

\subsection{Motion Intensity Resampling}
To balance the training dataset, we resampled the videos based on their motion intensity values, measured within a range of 0 to 20. Specifically, we adjusted the sampling to ensure that videos across different motion intensity levels are more evenly represented within this range, balancing the distribution of videos across varying degrees of motion. For videos with motion intensity values exceeding 20 (which constitute a minority within the dataset), we capped their motion intensity at 20. This approach creates a more balanced distribution of motion intensity levels across the dataset.

\subsection{The Human-Motion Dataset} The Human-Motion dataset consists of 106,292 video clips sourced from various datasets, including VFHQ~\cite{xie2022vfhq} (1,843 clips), CelebV-Text~\cite{yu2023celebv} (52,072 clips), CelebV-HQ~\cite{zhu2022celebv} (31,004 clips), AAHQ~\cite{liu2021blendgan} (17,619 clips), and a private dataset, Sing Videos (3,752 clips). Each clip was rigorously filtered and re-annotated to ensure high-quality identity and motion information across diverse formats, resolutions, and styles.

\section{Additional Visual Results}
\label{sec:vis}
\subsection{Action Phrase Control}
\begin{figure}[!t]
    \centering
    \includegraphics[width=\columnwidth]{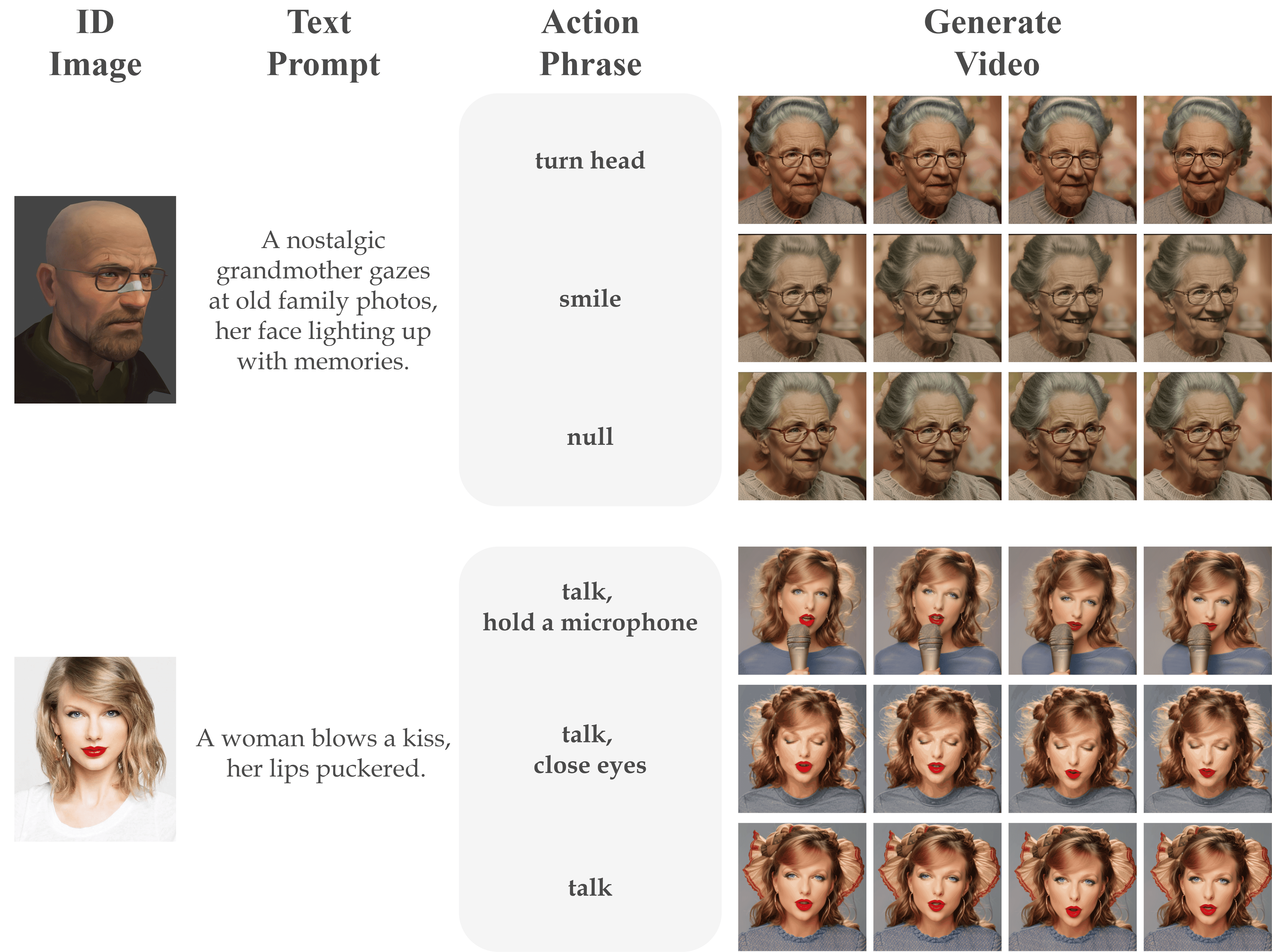}
    \caption{Action phrase control demonstration. Our model generates corresponding motions for different action phrases (e.g., ``turn head", ``smile", ``null", ``talk, hold a microphone") while maintaining consistent identity and context. The results showcase \textit{MotionCharacter}'s ability to handle both simple and compound action instructions.}
    \label{fig:motioncontrol}
\end{figure}

To evaluate our model's responsiveness to different action instructions, we conduct controlled experiments by varying only the action phrases while keeping the reference image, text prompt, and motion intensity constant. As shown in Fig.~\ref{fig:motioncontrol}, our model accurately interprets and executes diverse action phrases such as ``turn head", ``smile", and ``talk, hold a microphone". 

\begin{figure}[!t]
    \centering
    \includegraphics[width=\columnwidth]{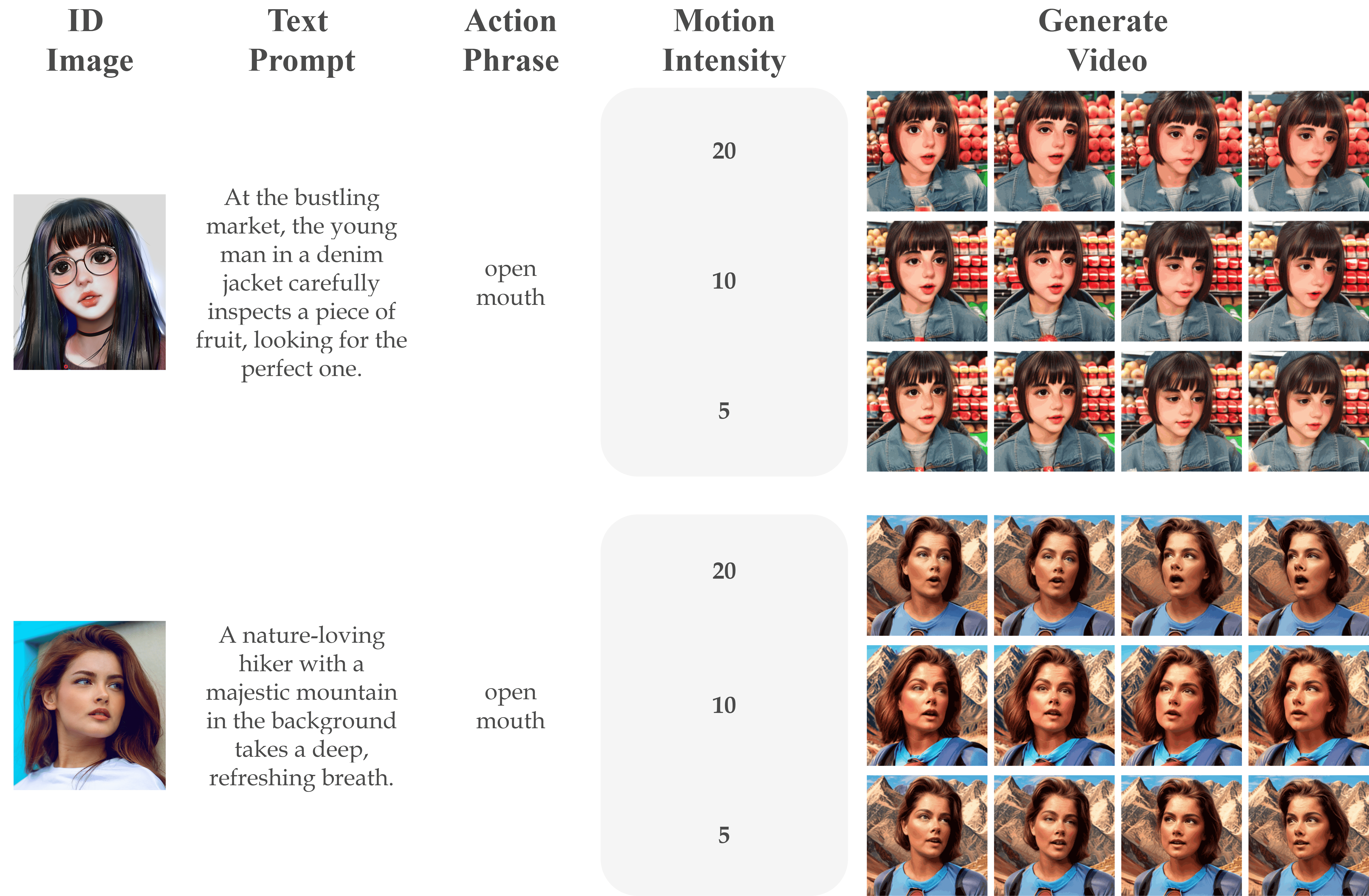}
    \caption{Motion intensity control demonstration. With fixed reference image and action phrase (``open mouth"), we show generated results at different intensity levels. Lower values (5) produce subtle movements, while higher values (20) generate more pronounced actions.}
    \label{fig:intensitycontrol}
\end{figure}

The results demonstrate two key capabilities: (1) precise action execution - each phrase generates distinct and appropriate movements, from subtle facial expressions to more complex combined actions; and (2) identity preservation - the subject's core features remain stable across different motions, validating our framework's ability to disentangle identity and motion attributes.

\subsection{Motion Intensity Control}
We analyze our model's ability to modulate motion intensity by fixing all parameters except the intensity value. Fig.~\ref{fig:intensitycontrol} shows the results of varying intensity levels (5, 10, 20) for the same action phrase ``open mouth".

The visualizations reveal smooth gradation of motion magnitude: lower intensity values (5) produce subtle movements while maintaining high temporal coherence, whereas higher values (20) generate more pronounced actions without compromising motion quality or identity consistency.

\section{Additional Implementation Details}
\label{sec:impl}
Our implementation is built upon a large-scale pre-trained Video Diffusion Model (VDM)~\cite{guo2023animatediff}.
All experiments were conducted using 8 NVIDIA A100 GPUs (80GB), with the training process taking approximately 24 hours. 
The re-annotation of the video and image datasets used in our experiments was performed in parallel on 32 NVIDIA V100 GPUs (32GB). The batch size was set to 2, with gradient accumulation over 20 steps to effectively simulate a larger batch size, allowing for efficient GPU memory usage and stable training. The training data consisted of diverse video clips, which were preprocessed to a resolution of $512\times512$ pixels, with 16 frames sampled per video at a frame rate of 4 frames per second. Face detection and alignment were performed using the InsightFace library~\cite{deng2019arcface} to ensure well-aligned facial regions for training. 
Data augmentation included random horizontal flipping, resizing, and center cropping to maintain consistent input dimensions. Furthermore, text and image dropout rates were set to 0.05, with a 50\% probability of dropping the CLIP embeddings $E_{clip}$. 
These strategies enhanced the model's robustness by simulating various levels of input occlusion and missing information. 
To optimize training efficiency, we applied mixed precision using fp16 to reduce memory consumption, along with gradient checkpointing. 
Additionally, xFormers memory-efficient attention~\cite{xFormers2022} was activated to further optimize memory usage. 
We used the AdamW~\cite{loshchilov2017decoupled} optimizer with a learning rate of $1\times10^{-5}$. Our model was trained for a total of 12,000 steps. During training, only the ID-Preserving Adapter and Motion Control Module are optimized, with all other model components kept frozen.

\section{Limitations}
\label{sec:limit}
While our framework achieves significant performance in identity-consistent and controllable Text-to-Video (T2V) generation, it has limitations in handling highly complex or intricate motion sequences, where fine-grained motion dynamics may not be captured effectively. Additionally, the framework’s performance is inherently dependent on the capabilities of the underlying T2V base model, which can limit the quality of generated videos. As T2V base models advance, our approach is designed with potential adaptability in mind; future iterations may leverage more powerful video foundation models, such as Wan series~\cite{wan2025wan}, to enhance generalization and video fidelity in increasingly demanding scenarios.

\end{document}